\documentclass[letterpaper, 10 pt, journal, twoside]{IEEEtran}
\usepackage{graphicx}
\usepackage{comment}
\usepackage{amsmath,amssymb} % define this before the line numbering.
\usepackage{times}
\usepackage{epsfig}

\usepackage{color}
\usepackage{multirow}
\usepackage{tabularx}
\usepackage{adjustbox}
\usepackage{subfig}
\usepackage{siunitx}
\usepackage{textcomp}
\usepackage{booktabs}
\usepackage[dvipsnames]{xcolor}

\usepackage{amsfonts}       % blackboard math symbols
\usepackage{nicefrac}       % compact symbols for 1/2, etc.

\usepackage{stmaryrd} % short arrow
\usepackage{kotex} % for korean comments

\usepackage{xspace}
\makeatletter
\DeclareRobustCommand\onedot{\futurelet\@let@token\@onedot}
\def\@onedot{\ifx\@let@token.\else.\null\fi\xspace}
 
\def\ie{\emph{i.e}\onedot}

\def\etal{\emph{et al}\onedot}
\makeatother

\newcommand{\CH}[1]{{#1}}

\newcommand{\figref}[1]{Fig.~\ref{#1}}

\newcommand{\tabref}[1]{Tab.~\ref{#1}}

\newcommand{\secref}[1]{Sec.~\ref{#1}}

\usepackage{lipsum}

\newcommand{\RNum}[1]{\uppercase\expandafter{\romannumeral #1\relax}}

\begin{document}

\title{\LARGE \bf
Category-Level Metric Scale Object Shape and Pose Estimation}

\author{Taeyeop Lee$^{1}$, Byeong-Uk Lee$^{1}$, Myungchul Kim$^{1}$, and In So Kweon $^{1}$% 
\thanks{
Manuscript received: April 30, 2021; Revised July 27, 2021; Accepted August, 23, 2021.
This paper was recommended for publication by Editor Cesar Cadena Lerma upon evaluation of the Associate Editor and 
Reviewers’ comments.
This work was supported by the Technology Innovation Program (10070171, Development of core technology for advanced locomotion/manipulation based on high-speed/power robot platform and robot intelligence) funded by the Ministry of Trade, Industry and Energy (MOTIE, Korea). \textit{(Corresponding author: I. S. Kweon.)}
}
\thanks{$^{1}$T. Lee, BU Lee, M. Kim and I. S. Kweon are with the Robotics and Computer Vision Laboratory, KAIST, Daejeon, Republic of Korea (e-mail: taeyeop.trevor@kaist.ac.kr; byeonguk.lee@kaist.ac.kr; gritycda@kaist.ac.kr; iskweon77@kaist.ac.kr).
% {\tt \textbraceleft taeyeop.trevor, byeonguk.lee, gritycda, iskweon77\textbraceright @kaist.ac.kr}
}
\thanks{Digital Object Identifier (DOI): see top of this page.}
}

\markboth{IEEE ROBOTICS AND AUTOMATION LETTERS. PREPRINT VERSION. ACCEPTED August, 2021}
{Lee \MakeLowercase{\textit{et al.}}: Category-Level Metric Scale Object Shape and Pose Estimation}

\maketitle
% \thispagestyle{empty}
% \pagestyle{empty}

%%%%%%%%%%%%%%%%%%%%%%%%%%%%%%%%%%%%%%%%%%%%%%%%%%%%%%%%%%%%%%%%%%%%%%%%%%%%
%%%%%%%%%%%%%%%%%%%%%%%%%%%%%%%% ABSTRACT %%%%%%%%%%%%%%%%%%%%%%%%%%%%%%%%%%
%%%%%%%%%%%%%%%%%%%%%%%%%%%%%%%%%%%%%%%%%%%%%%%%%%%%%%%%%%%%%%%%%%%%%%%%%%%%
\begin{abstract}
\label{abstract}
% This work addresses the challenging problem of estimating the unseen object’s metric scale 3D shape and 6D pose.
Advances in deep learning recognition have led to accurate object detection with 2D images.
However, these 2D perception methods are insufficient for complete 3D world information.
Concurrently, advanced 3D shape estimation approaches focus on the shape itself, without considering metric scale.
These methods cannot determine the accurate location and orientation of objects. To tackle this problem, we propose a framework that jointly estimates a metric scale shape and pose from a single RGB image.
Our framework has two branches: the Metric Scale Object Shape branch (MSOS) and the Normalized Object Coordinate Space branch (NOCS).
The MSOS branch estimates the metric scale shape observed in the camera coordinates.
The NOCS branch predicts the normalized object coordinate space (NOCS) map and performs similarity transformation with the rendered depth map from a predicted metric scale mesh to obtain 6d pose and size.
Additionally, we introduce the Normalized Object Center Estimation (NOCE) to estimate the geometrically aligned distance from the camera to the object center.
We validated our method on both synthetic and real-world datasets to evaluate category-level object pose and shape.
% To reduce the scale ambiguity by providing the one-pixel value of depth measurement, our method can estimate the accurate pose and size of objects compared to the full pixel value of depth measurements (RGB-D).

\end{abstract}
% \begin{IEEEkeywords}
% IEEE, IEEEtran, journal, \LaTeX, paper, template.
% \end{IEEEkeywords}

% \fi
\IEEEpeerreviewmaketitle

%%%%%%%%%%%%%%%%%%%%%%%%%%%%%%%%%%%%%%%%%%%%%%%%%%%%%%%%%%%%%%%%%%%%%%%%%%%%
%%%%%%%%%%%%%%%%%%%%%%%%%%%%%%%% INTRODUCTION %%%%%%%%%%%%%%%%%%%%%%%%%%%%%%
%%%%%%%%%%%%%%%%%%%%%%%%%%%%%%%%%%%%%%%%%%%%%%%%%%%%%%%%%%%%%%%%%%%%%%%%%%%%
\section{INTRODUCTION}
\label{introduction}
% 6D Pose + Object Shape의 필요성 
3D object recognition is one of the crucial tasks in various robotics and computer vision applications, such as robot manipulation and augmented reality (AR).
To recognize and interact with objects in the 3D space, these applications require acquiring the types, shapes, sizes, locations, and orientations of the objects.
For example, for a robot to grasp a mug, it needs to know what kind of a mug it is (type), how it is shaped (shape), how big or small the mug is (size), where the mug is exactly lying (location), and in what direction the handle of the mug is pointing (orientation).
In augmented reality, these kinds of information also enable virtual interaction with the objects re-rendered from the real world.
The shape and size information can be referred to as a metric scale object shape.
The location and orientation of an object can be interpreted as a 6D object pose, with 3 degrees of freedom for each.

% 기존의 RGB + Model Based의 방법론 (Instance level)
Many of the previous 6D object pose estimation methods have tried to solve instance-level pose estimation, by assuming that the 3D CAD model and size information of each object is known.
Recent RGB-based approaches~\cite{peng2019pvnet, tekin2018real_indirect} have shown remarkable performances utilizing the Perspective-n-Point (PnP) algorithm \cite{lepetit2009epnp}.
However, it is not easy to have or obtain accurate 3D scans of every object, and this makes it difficult to apply these algorithms in general settings where the objects have never been seen before, and therefore have no 3D CAD models.
Also, it is inefficient to train a separate network each time a new object is observed.

% 기존의 RGB-D + Model Free의 방법론 (Category level)
To address this problem, category-level 6D pose and size estimation approaches have been proposed to handle unseen objects.
Category-level pose estimation tasks assume that the object instances have not been previously seen, but their categories are still within known categories. 
Wang~\etal~\cite{wang2019normalized} introduced a new representation called Normalized Object Coordinate Space (NOCS), to align different object instances within one category in a shared orientation.
By predicting a category-wise similar NOCS map, it can estimate the 6D pose of an unseen object without a 3D model.
To estimate an object pose and size information, previous category-level 6D pose estimation methods~\cite{wang2019normalized, Tian2020prior, chen2020cass} have required depth information instead of 3D model information.
Chen~\etal~\cite{chen2020Synthesis} proposed an algorithm that predicts the shape and the pose of an object without using depth information, but its RGB-based object shape estimation results lack object size information. 
Moreover, current category-level pose estimation methods do not deal with metric scale object shapes in 3D space.

\begin{figure}
\begin{center}
\includegraphics[width=1.0\linewidth]{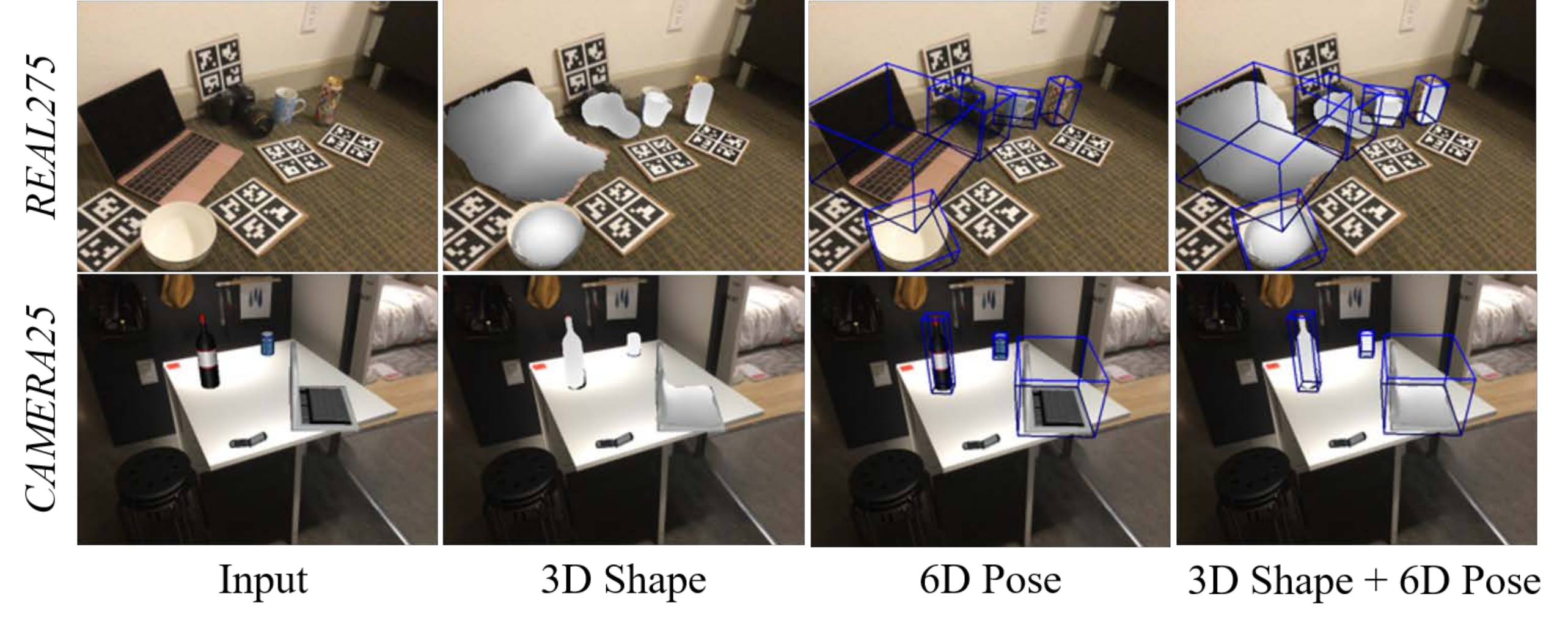} \\
% \vspace{1.0mm}
\caption{\textbf{3D shape and 6D Pose estimation examples of our proposed framework.} Our framework takes an input image and predicts the metric scale 3D shape, 6D pose of the object.
}\label{fig:main}
\end{center}
\vspace{-5mm}
\end{figure}

\begin{figure*}
\begin{center}
\includegraphics[width=0.8\linewidth]{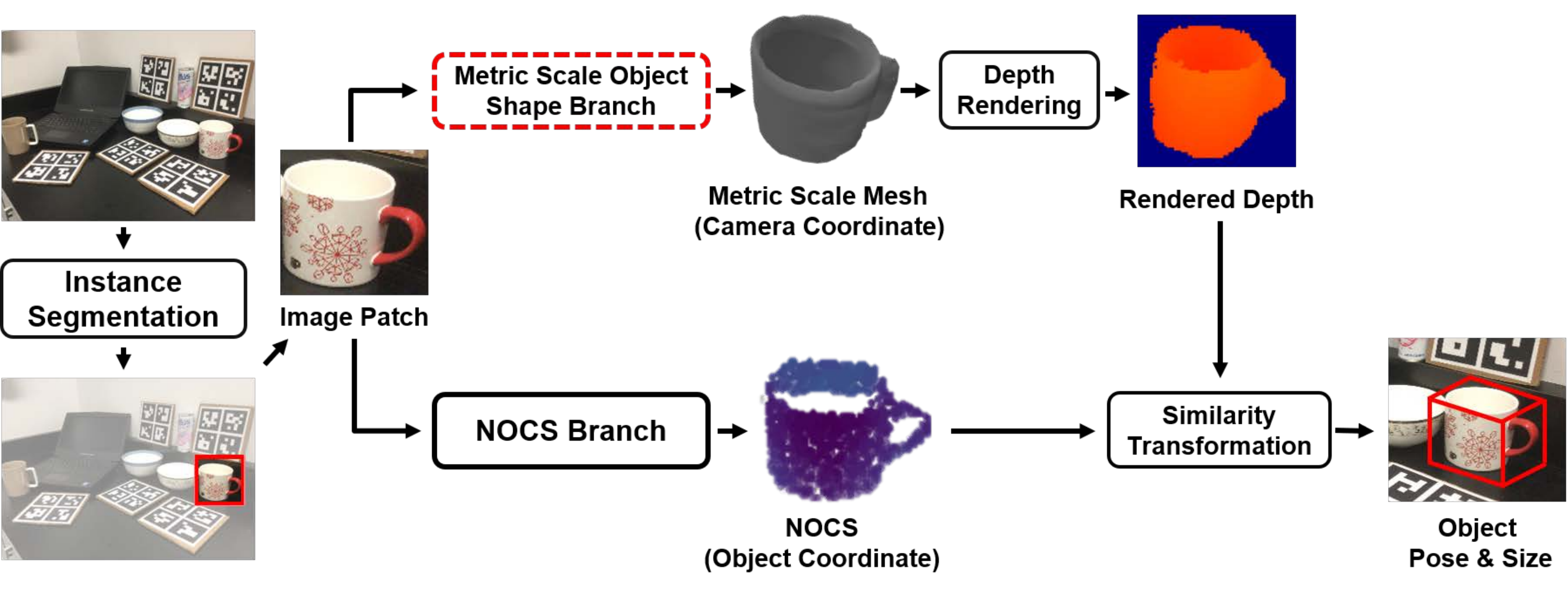} \\
% \vspace{1.0mm}
\caption{\textbf{An overview of our framework.} Our framework first detects class label, bounding box and segmentation mask using the off-the-shelf instance segmentation method. The detected image patch goes into two branches: a metric scale object shape branch (MSOS) and a normalized object coordinate space branch (NOCS).
The MSOS branch estimates the shape of the metric scale objects and rendered depths by projecting the predicted metric scale object mesh onto the image plane.
The 6D object pose is obtained by similarity transformation using a rendered depth map and the predicted NOCS map.
}
\label{fig:framework}
\end{center}
\vspace{-5mm}
\end{figure*}

In this paper, we propose a CNN-based category-level metric scale object shape and pose estimation algorithm, utilizing only a single RGB image with none or very sparse depth information.
Our algorithm consists of two neural network branches, where one performs the metric scale mesh estimation and the other one estimates the NOCS map.
We replace the 3D CAD model or the depth information that was used in the previous 6D pose estimation approaches, with the metric scale mesh itself and the rendered depth map acquired by projecting the mesh onto the image plane.
With the obtained NOCS map and the rendered depth map, our network predicts the size and 6D pose of an object via similarity transformation~\cite{umeyama1991least}.
Our category-level object shape, pose and size estimation algorithm had the best performances among RGB-based 6D pose estimation methods in the benchmark evaluation.
Also, without any additional training, our algorithm refines the scale and size prediction using only a few sparse depth inputs, and showed results comparable to the previous RGBD-based approaches in object size and pose estimation.
We additionally performed various ablation studies to show the effectiveness and validity of our design choices.

%%%%%%%%%%%%%%%%%%%%%%%%%%%%%%%%%%%%%%%%%%%%%%%%%%%%%%%%%%%%%%%%%%%%%%%%%%%%%%%%
%%%%%%%%%%%%%%%%%%%%%%%%%%%%%%%% RELATED WORK %%%%%%%%%%%%%%%%%%%%%%%%%%%%%%%%%%
%%%%%%%%%%%%%%%%%%%%%%%%%%%%%%%%%%%%%%%%%%%%%%%%%%%%%%%%%%%%%%%%%%%%%%%%%%%%%%%%

\begin{figure*}
\begin{center}
\includegraphics[width=0.8\linewidth]{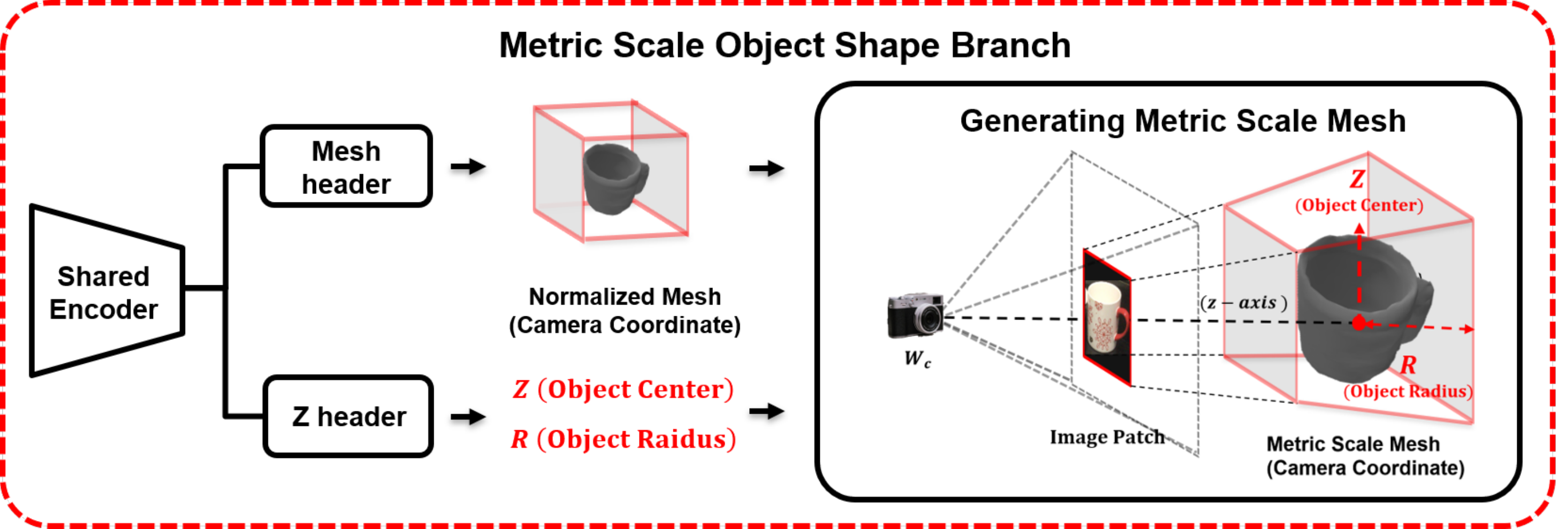}
\caption{\textbf{Illustration of the MSOS branch.} 
Our metric scale object shape branch (MSOS) has two headers with a shared feature encoder.
The mesh header estimates a normalized object mesh.
The Z header estimates the object center and radius to lift the normalized object mesh to the metric scale object mesh. Graph convolution is used to refine the mesh.
}
\label{fig:metric_object_shape_branch}
\end{center}
\vspace{-5mm}
\end{figure*}

\section{RELATED WORK}
\subsection{Instance-level object pose estimation}
With the increasing interest in 3D object recognition, various instance-level object pose estimation approaches have been introduced.
Instance-level pose estimation approaches can be broadly divided into two categories: correspondence-based methods and regression-based methods. 
The regression-based methods directly infer the rotation and translation from input images.
The challenging part of these approaches is that non-linearity in the pose space makes training hard to generalize \cite{sattler2019understanding}.
To get an accurate pose, these methods require certain types of refinement processes such as ICP which are based on 3D CAD models~\cite{rusinkiewicz2001efficient}.
Correspondence-based methods rely on point matching between the 2D image and a 3D model.
\cite{park2019pix2pose, zakharov2019dpod, li2019cdpn} regressed the corresponding 3D object coordinates for each image pixel in the masked region.
\cite{peng2019pvnet, rad2017bb8_indirect, tekin2018real_indirect, tremblay2018deep} detected the keypoints in the projected 3D points on the image and then solved a non-linear problem using the Perspective-n-Point (PnP) algorithm~\cite{lepetit2009epnp}.
While all of the aforementioned instance-level works have shown impressive results, most of them require 3D CAD model information at both the training and inference times. 
This dependency on 3D CAD models limits their applicability since storing and updating all objects at test time is time consuming and impractical for real-world scenarios.

\subsection{Category-level object pose estimation}
Some methods have focused on estimating the pose of unseen objects using known categories during training when no instance-specific 3D models are given at test time. 
As a pioneering work, Wang \etal \cite{wang2019normalized} introduced a Normalized Object Coordinate Space (NOCS) for category-level 6D object pose estimation. 
Wang \etal estimated NOCS, the shared canonical representation of object instances within a category, from the RGB image.
It indirectly estimates the pose and size of an object by matching their predicted NOCS map and observed depth map using non-linear solutions \cite{umeyama1991least}.
Tian \etal \cite{Tian2020prior} proposed a more advanced network that explicitly learns deformation from the shape prior to estimating the NOCS map.
Chen \etal \cite{chen2020cass} regressed the pose and size directly from the image and depth.
However, to reliably predict object size, these methods inevitably require depth information.
Recently, Chen \etal \cite{chen2020Synthesis} proposed a generative model that generates the object appearance from various viewpoints.
Synthesis \cite{chen2020Synthesis} estimates an object pose from an RGB image by matching the input and the generated appearance using an iterative alignment at inference time.
Therefore, this approach does not handle 3D object shapes and sizes.

%%%%%%%%%%%%%%%%%%%%%%%%%%%%%%%%%%%%%%%%%%%%%%%%%%%%%%%%%%%%%%%%%%%%%%%%%%%%%%%%
%%%%%%%%%%%%%%%%%%%%%%%%%%%%%%%% METHOD %%%%%%%%%%%%%%%%%%%%%%%%%%%%%%%%%%%%%%%%
%%%%%%%%%%%%%%%%%%%%%%%%%%%%%%%%%%%%%%%%%%%%%%%%%%%%%%%%%%%%%%%%%%%%%%%%%%%%%%%%

% Methods
\section{Category-Level Metric Scale Object Shape and Pose Estimation}
% 전반적인 Task와 설명
Given an RGB image, our goal is to estimate the object shape, pose, and size of an object instance.
\figref{fig:framework} shows the overall pipeline of our framework.
Our framework detects the class label, bounding box, and segmentation using an off-the-shelf instance segmentation network.
The detected image patches are then used as inputs of the two branches: a metric scale object shape branch (MSOS) and a normalized object coordinate space branch (NOCS).
% depth를 replacing 하는 방법 + pose를 추정하는 방법 설명 
The MSOS branch estimates the object shape in the metric scale (\secref{subsec:metric_scale_object_shape}) and renders 2D depth maps by projecting the predicted metric scale object shape onto the image plane (\secref{subsec:depth_rendering}).
The rendered depth map and predicted NOCS map from the NOCS branch are used to estimate the pose and size of the object using the similarity transformation (\secref{subsec:pose_estimation}).

\begin{figure}
\begin{center}
\includegraphics[width=0.7\linewidth]{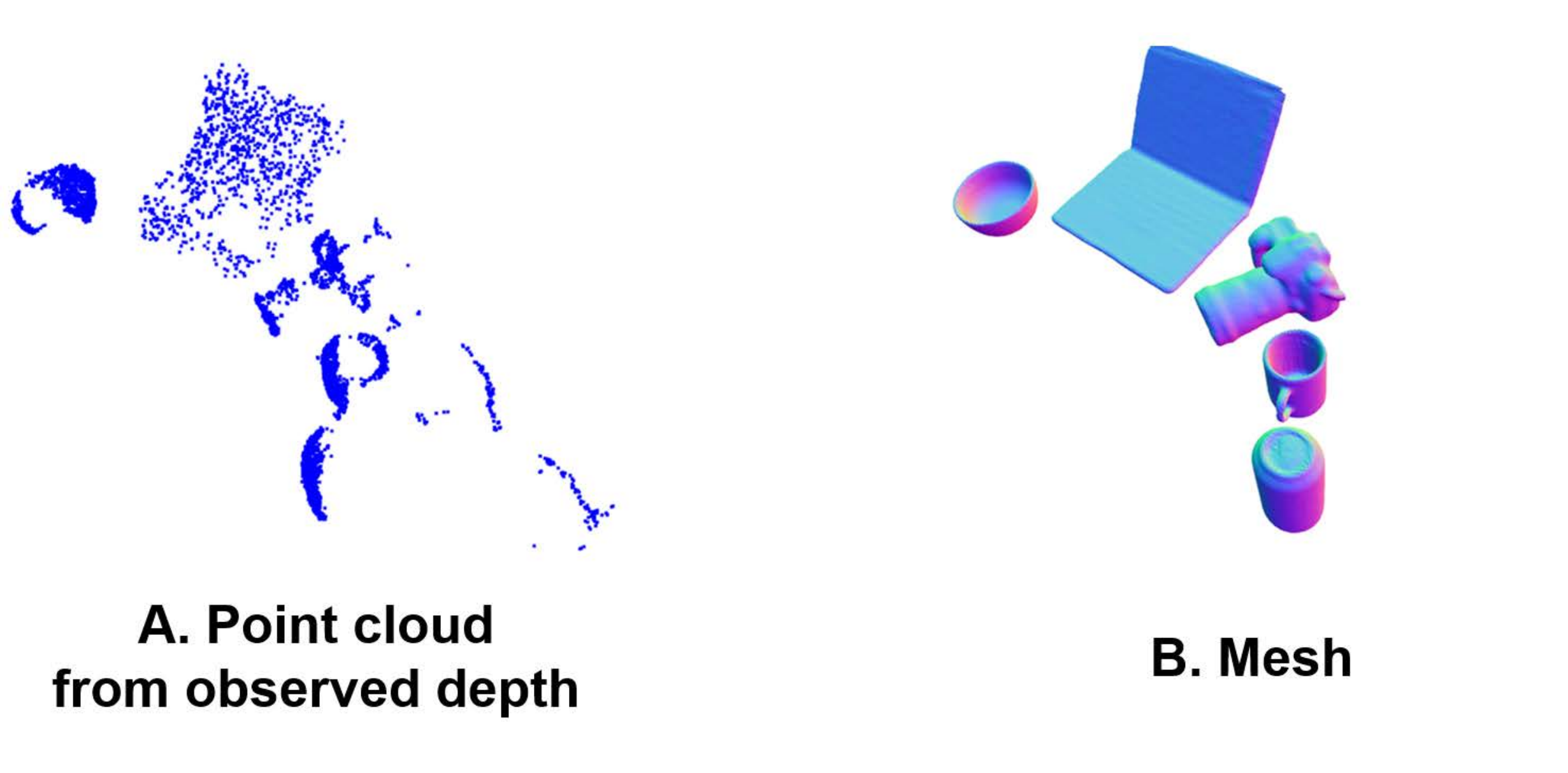} \\
\vspace{1.0mm}
\caption{\textbf{Difference between point cloud and mesh representation.} 
Compared to mesh representation, point cloud representation does not contain information on the object surface and invisible parts, therefore the shape is too sparse.
}
\label{fig:point_mesh}
\end{center}
\vspace{-5mm}
\end{figure}

\subsection{Metric Scale Object Shape branch (MSOS)}
\label{subsec:metric_scale_object_shape}
Various methods have been developed for different 3D shape representations, such as voxels \cite{choy20163d, brock2016generative,wu2016learning}, point clouds \cite{kurenkov2018deformnet, fan2017point, mandikal20183d} and meshes \cite{gkioxari2019mesh, groueix2018papier, pan2019deep, wang2018pixel2mesh}.
In robotics and AR applications, we believe that dense and accurate 3D shape representation is necessary.
The accuracy of the detailed structure in a voxel representation is dependent on the voxel resolution in 3D space.
Point cloud representation may handle more details with less-quantized 3D points, but as shown in \figref{fig:point_mesh}, it does not contain object surface information and as a result, the shape is too sparse.
For these reasons, we chose the mesh representation, which is capable of showing both dense and accurate 3D shapes.
Also, mesh enables the depth rendering process described in \secref{subsec:depth_rendering}.

The Metric Scale Object Shape (MSOS) branch estimates the metric scale object mesh from the camera coordinate using a single image.
\figref{fig:metric_object_shape_branch} shows details of the MSOS branch.
The MSOS branch consists of two jointly-trained headers with a shared encoder.
The mesh header estimates the normalized object mesh, and the Z header predicts $Z$ information, which is the distance of the object along the z-axis from the camera center, to lift the normalized object shape to a metric scale object shape.

% Mesh RCNN을 사용한 이유
\subsubsection{Normalized Object Mesh Estimation (Mesh header)}
\label{subsec:mesh_header}
Instead of training a single network for each category \cite{peng2019pvnet, rad2017bb8_indirect, tekin2018real_indirect, tremblay2018deep}, we aim to cover all object categories with a single network~\cite{Tian2020prior, wang2019normalized}.
For this reason, while not being sensitive to any specific network baseline, we follow a similar scheme as the Mesh R-CNN \cite{gkioxari2019mesh} to estimate the normalized object mesh $M_{\text{norm}}$.
% Mesh RCNN의 과정 간략히 설명
The detected object in the instance segmentation stage is cropped using the bounding box and is resized to have the same patch size.
By taking this resized image patch as input, the mesh header can efficiently learn the object shape by utilizing just the relevant information.

% Z header 설명
\subsubsection{Normalized Object Center Estimation (NOCE)}
\label{subsec:noce}
To transform the predicted normalized mesh into the camera coordinate, \ie metric scale, we need the location and scale information of the object.
As shown in \figref{fig:metric_object_shape_branch}, we represent the distance from the camera to the object center in the z-axis as object center $Z$, and we also define the radius of the metric scale object mesh as object radius $R$.
The object center handles the location of the metric mesh, and the object radius decides the scale/size.
We call our object center and object radius estimation branch the Z header since both values are predicted along the z-axis from the camera.

Estimating the distance to the center of an object using only an RGB image is challenging.
The main reason is that the RGB image does not contain any information regarding 3D volume.
To tackle this problem, we implicitly embed the mesh prediction feature in the training process, using a shared feature encoder for the mesh header and Z header.
We believe that by jointly estimating object shape and object center, each header can be guided using the objective of the other header.

\begin{figure}
\begin{center}
\includegraphics[width=0.7\linewidth]{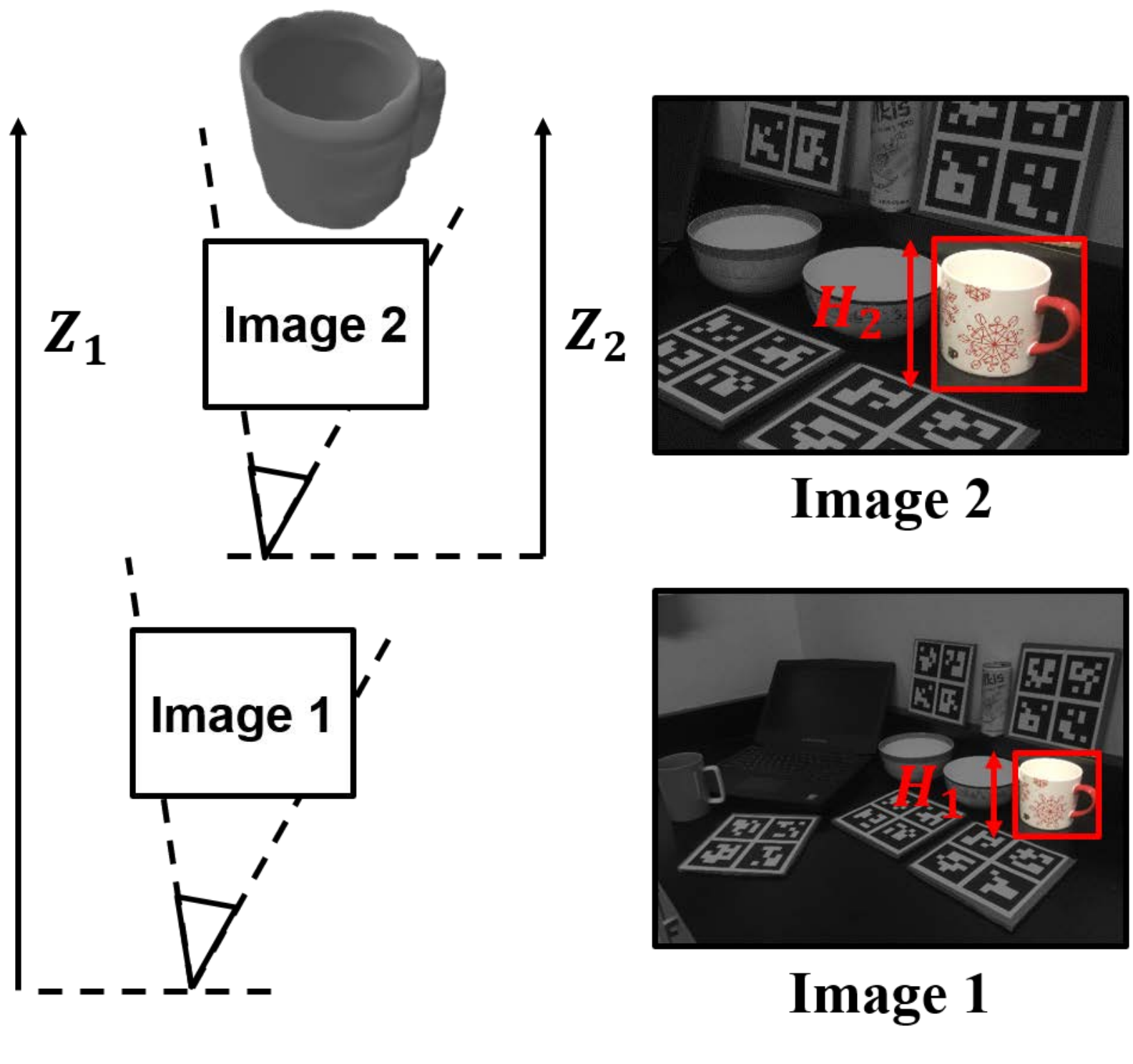} \\
\caption{\textbf{Relationship between the size of the bounding box and the distance from the camera center.}
}
\label{fig:NOCE}
\end{center}
\vspace{-5mm}
\end{figure}

As explained in~\secref{subsec:mesh_header}, using a cropped and resized image on the object region as input for our mesh header is beneficial for mesh quality and efficient inference.
However, when resizing the cropped image patch, the information in the detected bounding box is lost.
This is crucial since the size of the bounding box implies the distance from the camera to the object, Accordingly, resizing the image will cause ambiguity for on $Z$ and $R$ prediction.
For example, as illustrated in~\figref{fig:NOCE}, upsampling the original image patch is similar to taking the camera closer to the scene, which alters the original $Z$ information.
Therefore, the two detected mugs appear to be the same size in the input image patch, while having a completely different object center.
This makes the object center prediction using a cropped and resized image patch an ill-posed problem.

To solve this problem, we propose the normalized object center estimation (NOCE).
The NOCE linearly scales the ground truth object center $Z$ by using the size information of the detected bounding box.
We compensate the resizing ratio $\tau$ to the object center $Z$, as
\begin{equation}
    Z_{\text{NOCE}} = \frac{Z \times \tau}{\CH{f}},
\end{equation}
where $\tau = H_o/H_{\text{patch}}$ while $H_o$ refers to the size of the original bounding box, $H_{\text{patch}}$ is the size of the network input image patch, and \CH{$f$ denotes the camera focal length}.
Instead of the original $Z$, the normalized object center $Z_{\text{NOCE}}$ is used as ground truth during training.
By doing this, similar objects in similar scenes will have similar $Z_{\text{NOCE}}$ values, which significantly helps to improve the metric scale object shape and translation estimation.
At inference time, we divide the predicted $Z$ by $\tau$ to restore the original object center information.

The object radius $R$ is also jointly learned by our Z header.
In the same context as NOCE, the object radius $R$ is trained on the normalized scale.
The final prediction $R$ is then multiplied by $Z$ to restore the original object size.

\subsubsection{Generating the Metric Scale Mesh}
\label{subsec:Z_header}
The final stage of our MSOS branch is to convert the predicted normalized mesh from the mesh header to metric scale mesh, using object center $Z$ and object radius $R$ estimation values from the Z header, as in,
\begin{equation}
    M_{\text{metric}} = f_{\text{scale}} (K, M_{\text{norm}}, Z, R).
\end{equation}
$f_{\text{scale}}$ transforms the normalized mesh to camera coordinate systems to make metric scale objects with the camera intrinsic $K$.
\figref{fig:metric_object_shape_branch} shows the intuitive process of $f_{\text{scale}}$ and how it lifts the normalized object shape to metric scale object shape from a detected image patch.
Using the original 2D bounding box, the predicted $Z$ and $R$, $f_{\text{scale}}$ obtains a 3D bounding box on the camera coordinate.
It then optimizes the normalized mesh into metric scale using keypoint correspondences between the normalized 3D bounding box and the estimated 3D bounding box.

\begin{figure}
\begin{center}
\includegraphics[width=1.0\linewidth]{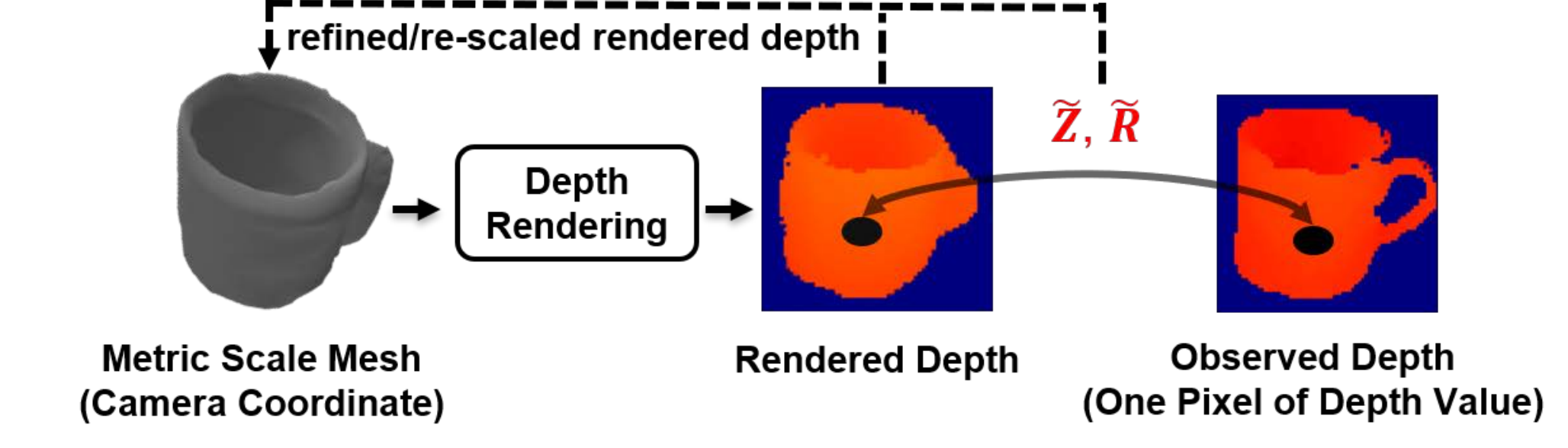} 
\caption{\textbf{Illustration of the depth rendering and refinement process.} 
The MSOS branch can additionally get a rendered depth map by projecting the metric scale mesh to the image plane. 
The rendered depth map and metric scale object mesh improve when very sparse depth is given by refining/re-scaling the object center and radius.
}
\label{fig:metric_object_shape_refinement}
\end{center}
\vspace{-5mm}
\end{figure}

\subsection{Depth Rendering}
\label{subsec:depth_rendering}
Previous NOCS-based approaches~\cite{wang2019normalized, Tian2020prior} use the predicted NOCS map and observed depth map to estimate the pose by similarity transformation.
One of the naive approaches to replace the observed depth information in an RGB setup is to apply monocular depth estimation.
However, we choose to project our metric scale object mesh into the image plane to obtain a rendered depth map.
We believe that our metric scale object shape generates more qualified depth for pose estimation than directly regressing the depths.
We use the DIB-R renderer~\cite{chen2019learning, wang2020self6d}, camera intrinsic $K$, and our predicted metric mesh $M_{\text{metric}}$ for our depth rendering.
The rendered depth map is denoted as $D_r$.
Directly regressing the depth information for each pixel makes it difficult to consider the local surface connection and overall 3D structure.
Our depth, rendered through the metric scale object surface, makes it possible to estimate the dense depth with each pixel relation using faces.
We compared the metric scale based depth and the direct regression depth in an ablation study (\secref{subsec:ablation_study}).

% sparse depth가 필요한 환경
Inspired by Cartman, winner of the 2017 Amazon robotics challenge \cite{morrison2018cartman}, we considered a setup where very sparse depth information is given in the depth rendering process.
Note that our method is RGB-based, but can refine the rendered depth map using sparse depth input without additional training.
When the additional sparse depth map is given, we can correct the initially predicted $Z$ and $R$ to $\tilde{Z}$ and $\tilde{R}$, using the difference between the observed sparse depth and the depth value of the matching pixel from our rendered depth map $D_r$.
Then, we regenerate the metric scale object mesh $\tilde{M}_{\text{metric}}$ with the normalized mesh $M_{\text{norm}}$, $\tilde{Z}$, and $\tilde{R}$.
Using $\tilde{M}_{\text{metric}}$ and the DIB-R renderer mentioned before, we get the refined/re-scaled rendered depth map $\tilde{D}_r$.
Specifically, we experimented on a setup where we used the depth value of only one pixel per object. We represent this setup as RGB-OD.
\figref{fig:metric_object_shape_refinement} shows the modifying process.

\subsection{Pose Estimation}
\label{subsec:pose_estimation}
% et al로 표현하는지 or 그냥 Wang으로 표현해도 되는지??
In the category-level pose estimation, Wang \etal \cite{wang2019normalized} and Tian \etal \cite{Tian2020prior} showed effective pose and size estimation using a normalized object coordinate space (NOCS) map.
The NOCS has the advantage of representing different object instances in the same category to the unified object coordinate space.
Existing NOCS based methods \cite{wang2019normalized, Tian2020prior} estimate the \CH{similarity transformation parameters (rotation, translation, and size)} by matching the pixel location of observed depth $D_o$ and the predicted NOCS map.
Compared to these approaches, we used the depth map $D_r$ rendered from our predicted metric scale mesh in the MSOS branch.
Given the rendered depth $D_r$ and NOCS $N_{c}$, \CH{the object rotation, translation} and size are estimated via the Umeyama algorithm \cite{umeyama1991least} and RANSAC \cite{fischler1981random}.

\subsection{Loss}
\label{subsec:loss_function}
To train our framework, we introduced our loss functions in the MSOS branch.
\CH{
The MSOS consists of two main headers: the mesh header and Z-header.
To train the mesh header, we adopted voxel loss $\zeta_\text{voxel}$ and normalized mesh loss $\zeta_{M_\text{norm}}$ from Mesh R-CNN \cite{gkioxari2019mesh}, where voxel loss computes the binary cross entropy (BCE), measured on a quantized voxel grid, and the normalized mesh loss is a combination of the chamfer distance between the predicted and ground truth point clouds, the inner product between the predicted and ground truth surface normal of the mesh, and the shape-regularizing edge loss.
For Z header, we applied radius loss $\zeta_R$ and our proposed normalized object center estimation (NOCE) loss $\zeta_{Z_\text{NOCE}}$, which are simple $smooth$ $l_1$ losses.}

Finally, we defined the MSOS branch loss functions: 
\begin{equation}\label{eq:all_loss}
\begin{split}
 \zeta_{\text{MSOS}} =  \lambda_{\text{voxel}} \zeta_{\text{voxel}} + \lambda_{M_{\text{norm}}} \zeta_{M_{\text{norm}}} \\
 + \lambda_{R} \zeta_{R} + \lambda_{Z_{\text{NOCE}}} \zeta_{Z_{\text{NOCE}}},
 \end{split}
\end{equation}
\CH{
where $\lambda_{\text{voxel}}$, $\lambda_{M_{\text{norm}}}$, $\lambda_R$ and $\lambda_{Z_{\text{NOCE}}}$ are weighting parameters, which are set empirically.}

%%%%%%%%%%%%%%%%%%%%%%%%%%%%%%%%%%%%%%%%%%%%%%%%%%%%%%%%%%%%%%%%%%%%%%%%%%%%%%%%
%%%%%%%%%%%%%%%%%%%%%%%%%%%%%%%% Experiments %%%%%%%%%%%%%%%%%%%%%%%%%%%%%%%%%%%%
%%%%%%%%%%%%%%%%%%%%%%%%%%%%%%%%%%%%%%%%%%%%%%%%%%%%%%%%%%%%%%%%%%%%%%%%%%%%%%%%
% Experiments
\section{Experiments}
\noindent{\textbf{Datasets.}}
We evaluated our method on two standard benchmarks in the task of category-level object pose estimation: Context-Aware MixedEd ReAlity (CAMERA) and the REAL dataset \cite{wang2019normalized}.
The CAMERA dataset is generated by rendering and compositing synthetic objects into real images in a context-aware manner. 
It comprises 300k synthetic images, where 25K images are for the evaluation. 
The synthetic training set contains 1085 object instances selected from 6 different categories - bottle, bowl, camera, can, laptop and mug. 
The evaluation set contains 184 different instances. 
The REAL \cite{wang2019normalized} dataset is complementary to the synthetic data. 
It has 43,000 real-world images of 7 scenes for training and 2,750 real-world images of 6 scenes for evaluation. 
It contains 42 unique objects with 6 categories. %The two evaluation sets are referred to as CAMERA and REAL.
Each evaluation set is noted as CAMERA25 and REAL275.

\noindent{\textbf{Metrics.}}
We followed the previous pose and size evaluation metric from the NOCS \cite{wang2019normalized}, which evaluates the performance of 3D object detection and 6D pose estimation.
We report the average precision at different Intersection-Over-Union (IoU) thresholds for 3D object detection.
Threshold values of 25\%, 50\%, and 75\% were used to evaluate the results.
For 6D object pose evaluation, the average precision was computed at rotation ($r^{\circ}$) and translation ($t$ cm) errors.
For example, the $10^{\circ}$ $10$cm metric denotes the average precision of object instances where the error is less than $10^{\circ}$ and $10$cm.

We used the chamfer distance ($\times 10^{-3}$) and normal consistency from \cite{gkioxari2019mesh} to evaluate our metric scale object shape.
To evaluate the depth, we adapted the standard evaluation metrics \cite{silberman2012indoor}. 
Let $\hat{D}_i$ denotes the ground truth depth at a pixel location $i$ and $D_i$ denotes the estimated depth.
The depth evaluation metrics are specified as follows,  $RMSE: \frac{1}{\left | I  \right |} \sum_{i \in {I}} \sqrt{(\hat{D}_i-D_i)^2}$, $REL: \frac{1}{\left | I  \right |} \sum_{i \in {I}} \left | (\hat{D}_i-D_i) / \hat{D}_i \right |$, \\$\delta_{\tau}:$  percentage of pixels satisfying $max (\frac{D_i}{\hat{D}_i},\frac{\hat{D}_i}{D_i}) < \tau$.

\begin{table*}\resizebox{\linewidth}{!}
    {
\begin{tabular}{ccc||cc|ccc|ccc|ccc}
\hline
\multirow{2}{*}{Depth} & \multirow{2}{*}{NOCE} & \multirow{2}{*}{Encoder} & \multicolumn{2}{c|}{3D Shape} & \multicolumn{3}{c|}{2D Depth} &\multicolumn{6}{c}{6D Pose and Size} \\ \cline{4-14}
& & & Chamfer ↓ & Normal ↓ & RMSE ↓ & Rel ↓ & $\delta_{1.25}$ ↑ &3D IoU (25) ↑ & 3D IoU (50) ↑ &  3D IoU (75) ↑ &  10 cm ↑ & $10^{\circ}$ ↑ & $10^{\circ}$ 10cm ↑ \\ \hline
Regression & & Sep. & - & - & 0.1700 & 0.2143 & 0.8274 & 50.1 & 15.4 &1.7 & 18.9 & 27.9 & 6.1 \\
Rendered & & Sep. & 40.08 & 0.76 & 0.1309 & 0.0984 & 0.9233 & 66.6 & 25.3 &4 & 24.8 & \textbf{63.8} & 16.6 \\
Rendered & \checkmark & Sep. & 34.34 & 0.82 & 0.1204 & 0.0914 & 0.9376 & 71.9 & 29.2 &4.6 & 28.0 & 61.6 & 18.2 \\ 
Rendered & \checkmark & Shared & \textbf{31.44} & \textbf{0.74} & \textbf{0.1168} & \textbf{0.0898} & \textbf{0.9486} & \textbf{75.4} & \textbf{32.4} & \textbf{5.1}& \textbf{29.7} & 60.8 & \textbf{19.2} \\ \hline
\end{tabular}
}
     \caption{
     \textbf{Ablation study of our design choices.} We compared 4 variations of our framework on the CAMERA25 dataset. We evaluated performances on 3D shape, 2D Depth, and 6D pose and size.
    }
    \label{tab:syn_ablation_study_NOCE_d}
% \vspace{-5mm}
\end{table*}

\noindent{}\textbf{Implementation details.}
Our MSOS branch uses a 4-channeled input, with an RGB image and a predicted segmentation mask, where we resize the input to 192 $\times$ 192 pixels.
We used the shared encoder structure from Chen \etal \cite{ChenPMVSNet2019ICCV}.
Due to the memory limitation, we set the 12 channels for voxel predictions.
In the Z header, each object center and radius decoders comprise 4 convolutional layers with kernel size 3 $\times$ 3 and 2 fully connected layers with ReLU activation.
We use the NOCS branch from off-the-shelf methods of NOCS predictor from \cite{Tian2020prior}.
Our overall pipeline was only trained on the CAMERA dataset.
We trained our MSOS branch from scratch, and we used the pretrained weights from the original paper for the NOCS branch.
We implemented our framework in PyTorch \cite{paszke2017automatic} and trained the MSOS branch using a single GPU (Titan Xp).
We empirically set the $\lambda_{\text{voxel}}$ = 3, $\lambda_{M_{\text{norm}}}$ = 1, $\lambda_{Z_{\text{NOCE}}}$ = 30, $\lambda_{R}$ = 30.
We used the Adam optimizer with an initial learning rate of 0.0001 and the batch size of 16.
In each stage, we decreased the learning rate by a factor of 0.6, 0.3, 0.1, 0.01 for each 15k, 30k, 45k, 60k iteration.

% Ablation Study
\subsection{Ablation study}
\label{subsec:ablation_study}
We performed an ablation study to evaluate our various design choices.
Both the training and testing were done on the CAMERA dataset.
Overall performances are summarized in \tabref{tab:syn_ablation_study_NOCE_d}.

\noindent{}\textbf{Depth rendering.} 
% depth rendering 
As mentioned in \secref{subsec:depth_rendering}, we compare our rendered depth map with simple UNet-based \cite{ronneberger2015u} monocular depth estimation results.
Comparing the first two rows in \tabref{tab:syn_ablation_study_NOCE_d}, it is clear that obtaining the depth map using our depth rendering with the predicted metric scale mesh outperformed all the metrics compared to the UNet-based depth estimation.
Also, UNet does not predict the metric scale shape of the 3D object, thereby the results are marked as `-' in the 3D shape metric.
In particular, the depth rendering results have a significant lead in the rotation evaluation.
We believe that our rendered depth contains the surface information between adjacent pixels based on face information from our predicted mesh.
These results demonstrate the superiority of our metric scale mesh based depth rendering approach over direct depth regression.

\begin{table*}
\centering
    \resizebox{0.8\linewidth}{!}{
    \begin{tabular}{c|c|ccc|ccc}
    \hline
    \multirow{2}{*}{Method} & \multirow{2}{*}{Input} & \multicolumn{6}{c}{6D Pose and Size} \\ \cline{3-8}
     & &3D IoU (25) ↑ & 3D IoU (50) ↑ &  3D IoU (75) ↑ &  10 cm ↑ & $10^{\circ}$ ↑ & $10^{\circ}$ 10cm ↑ \\ \hline
    Synthesis \cite{chen2020Synthesis} & RGB & - & - & - & 34.0 & 14.2 & 4.8 \\
    Ours & RGB & \textbf{62} & \textbf{23.4} & \textbf{3.0} & \textbf{39.5} & \textbf{29.2} & \textbf{9.6} \\ \hline
    NOCS \cite{wang2019normalized} & RGB-D & \textbf{84.4} & 76.9 & 30.1 & 97.9 & 24.8 & 24.3 \\
    Shape-Prior \cite{Tian2020prior} & RGB-D & 83.4 & 77.4 & \textbf{53.5} & 99.4 & \textbf{57.5} & 56.2 \\
    CASS \cite{chen2020cass} & RGB-D & \textbf{84.4} & \textbf{79.3} &  & & & \textbf{58.3} \\
    Ours & RGB-OD & 81.6 & 68.1 & 32.9 & 96.7 & 28.7 & 26.5 \\ \hline
    \end{tabular}    
    }
    % \vspace{1.5mm}
    \caption{\textbf{Quantitative comparison with the state-of-the-art methods on the REAL275 dataset.}
    Empty entries either could not be evaluated, or were not reported in the original paper.
    }
    \label{tab:real_rgbd_sota_compare}
% \vspace{-5mm}
\end{table*}

\begin{table*}
\centering
    \resizebox{0.8\linewidth}{!}{
    \begin{tabular}{c|c|ccc|ccc}
    \hline
    \multirow{2}{*}{Method} & \multirow{2}{*}{Input} & \multicolumn{6}{c}{6D Pose and Size} \\ \cline{3-8}
     & &3D IoU (25) ↑ & 3D IoU (50) ↑ &  3D IoU (75) ↑ &  10 cm ↑ & $10^{\circ}$ ↑ & $10^{\circ}$ 10cm ↑ \\ \hline
    Ours & RGB & \textbf{75.4} & \textbf{32.4} & 5.1 & \textbf{29.7} & 60.8 & \textbf{19.2} \\ \hline
    NOCS \cite{wang2019normalized} & RGB-D & 90.0 & 87.8 & 69.1 & 99.0 & 63.5 & 63.1 \\
    Shape-Prior \cite{Tian2020prior} & RGB-D & \textbf{94.2} & \textbf{93} & \textbf{84.5} & \textbf{99.4} & \textbf{80.1} &\textbf{79.8} \\
    Ours & RGB-OD & 93.8 & 90.8 & 72.8 & 96.6 & 60.7 &58.9 \\ \hline
    \end{tabular}    
    }
    \caption{\textbf{Quantitative comparison with state-of-the-art methods on the CAMERA25 dataset.}}
    \label{tab:syn_rgbd_sota_compare}
\vspace{-5mm}
\end{table*} 

\noindent{}\textbf{Object Center Estimation.}
As was explained in \secref{subsec:noce}, we proposed a shared feature encoder training and normalized object center estimation (NOCE) technique for robust object center estimation.
To ablate the effectiveness of our design choices, we used the same NOCS map in the evaluation.
The second row in \tabref{tab:syn_ablation_study_NOCE_d} does not use NOCE but directly estimates the object center $Z$ from an image patch that has not been resized.
\tabref{tab:syn_ablation_study_NOCE_d} shows that using NOCE and shared feature encoder improves their overall performance, respectively.
We want to focus more on the chamfer distance of the 3D shape evaluation and 3D IoUs and 10cm precision on the 6D pose and size evaluation, since these metrics are directly related to the object center $Z$ and the object radius $R$ prediction.
In these evaluation categories, it is shown that NOCE significantly improves the performances, since it enables robust object center prediction.
By implicitly guiding the training process with the shared encoder for our mesh header and our Z header, it also helps to boost the accuracy in the aforementioned metrics.
Overall, among our 4 variations of object shape and pose estimation approaches, all three of our network designs resulted in the best performance.

% Comparison with state-of-the-art
\subsection{Comparison with state-of-the-art}
\label{subsec:sota_comparison}

\begin{figure}
\begin{center}
\includegraphics[width=1.0\linewidth]{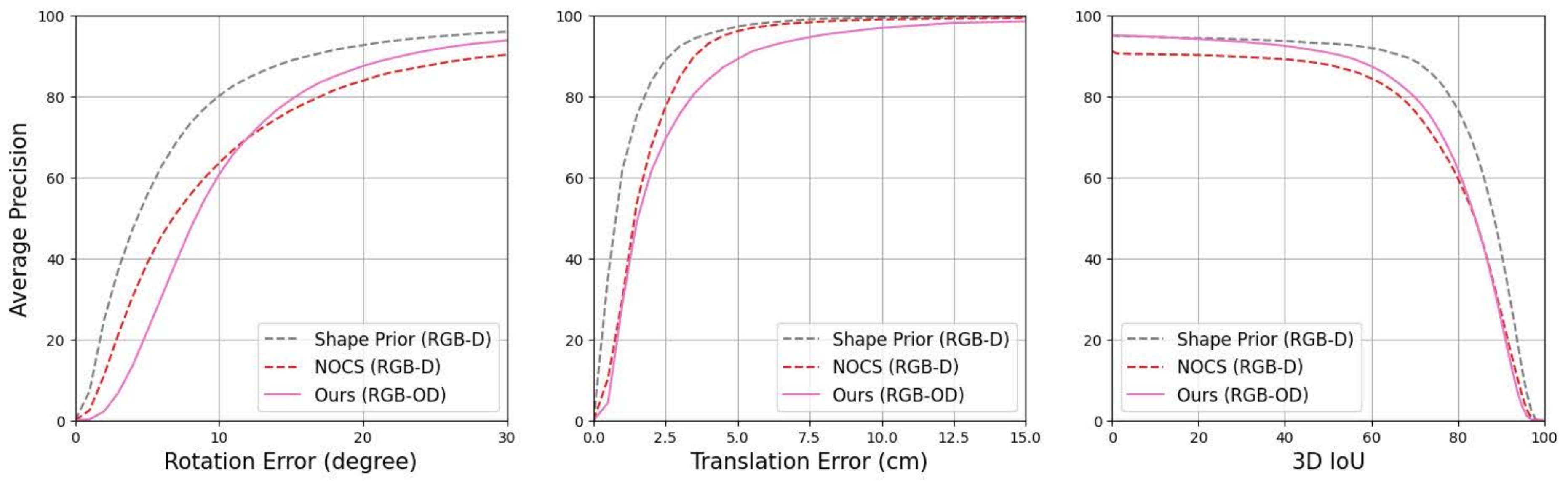} \\
\caption{\textbf{Comparison with state-of-the-art RGB-D methods on thef CAMERA25 dataset.}}\label{fig:syn_sota_rgbd_fig}
\end{center}
\vspace{-5mm}
\end{figure}

\subsubsection{RGB based methods}
\label{subsec:sota_comparison_rgb}
To the best of our knowledge, Synthesis by Chen~\etal~\cite{chen2020Synthesis} is the only method in the RGB-based category-level pose estimation.
Since Synthesis~\cite{chen2020Synthesis} has no results using the CAMERA25 dataset, we only compared our method on the REAL275 dataset.
Note that Synthesis~\cite{chen2020Synthesis} cannot estimate the size of an object. 
\tabref{tab:real_rgbd_sota_compare} shows that our method achieved state-of-the-art results in the RGB-based category-level pose estimation.
Especially, our method achieved more than double the performance at $10^{\circ}$ metric (29.2\% compared to 14.2\%).

\begin{figure*}
\begin{center}
\includegraphics[width=0.74\linewidth]{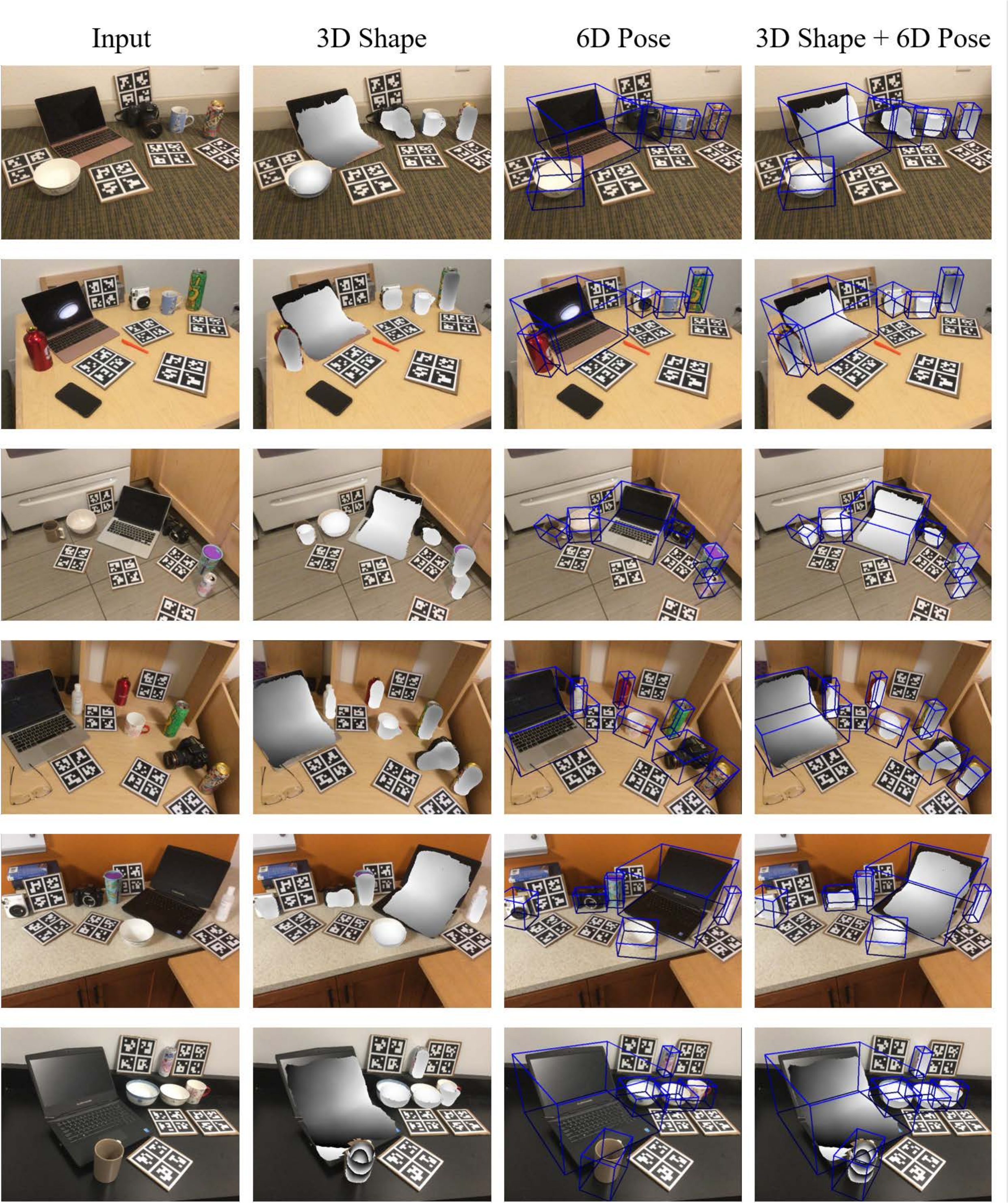} \\
\caption{\CH{\textbf{Qualitative results on the REAL275 dataset.}}}\label{fig:REAL275_qualitative}
\end{center}
\vspace{-5mm}
\end{figure*}

\begin{figure}
\begin{center}
\includegraphics[width=0.8\linewidth]{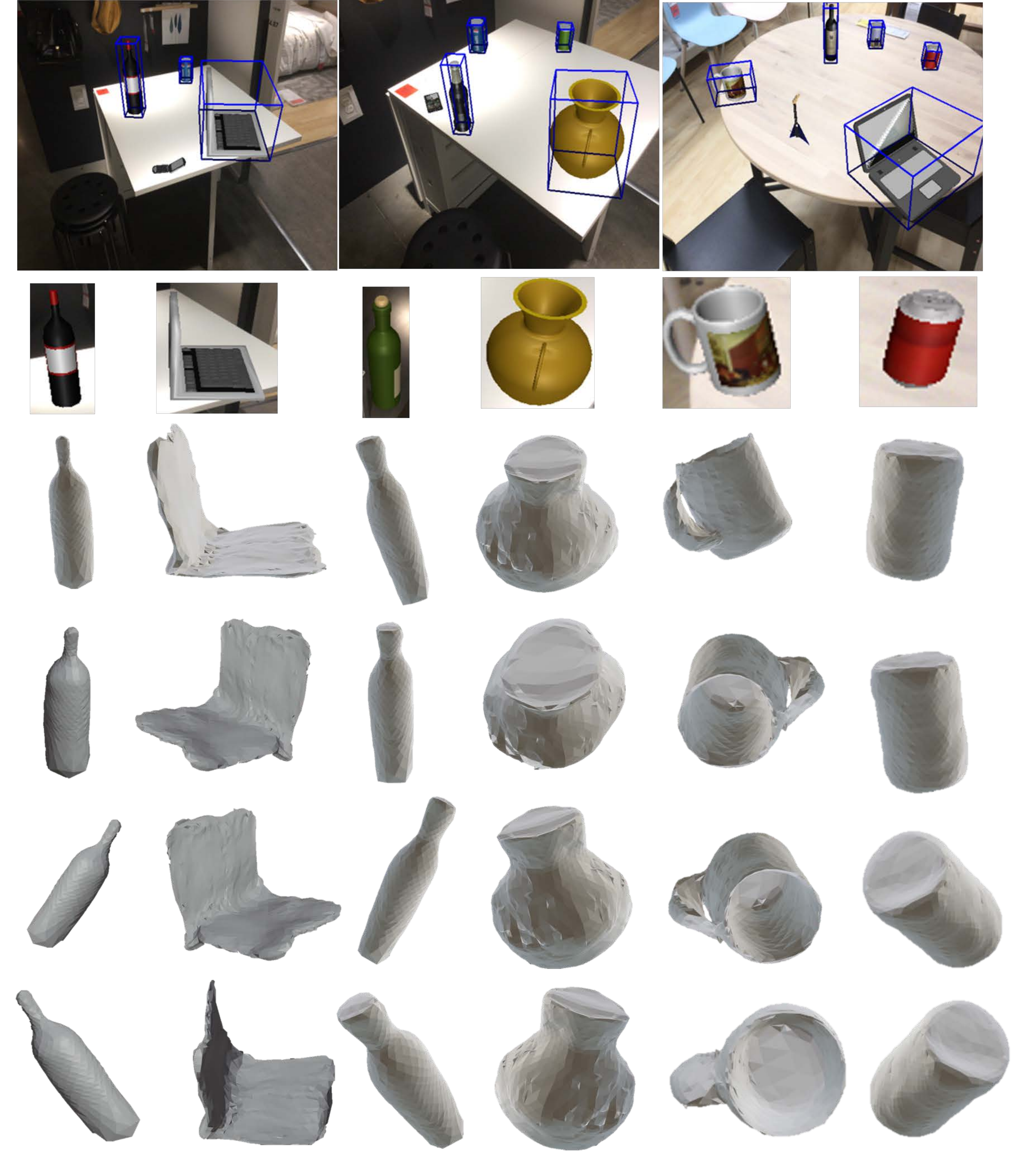} \\
\caption{\textbf{Qualitative examples of our predictions on CAMERA25.} 
 Our method jointly predicted the object shape with fine details, including bottle head, mug and thin laptop with accurate poses.
 Our metric scale object shape branch (MSOS) reconstructs the object mesh well even for invisible parts.
} 
\label{fig:qualitative}
\end{center}
\vspace{-5mm}
\end{figure}

\subsubsection{RGB-D based methods}
\label{subsec:sota_comparison_rgbd}
We also compared our performance on the RGB-OD setup with RGB-D-based state-of-the-art methods on both the CAMERA25 and REAL275 datasets.
As mentioned earlier, our RGB-OD setup uses only one pixel depth value per object.

The results on the REAL275 dataset are summarized in \tabref{tab:real_rgbd_sota_compare}, and the results on the CAMERA25 dataset are summarized in \tabref{tab:syn_rgbd_sota_compare}.
In the evaluation of the REAL275 dataset, note that unlike the existing RGB-D methods~\cite{wang2019normalized, Tian2020prior}, we trained only with the CAMERA dataset.
Both tables show that our RGB-OD results were significantly better than our results with the RGB-only setup.
While our RGB-based method still performed the best among RGB-only approaches, we can say that it is still a challenging problem to solve the object center estimation and object size estimation problem without using any reliable depth information.

Comparing our RGB-OD results to other RGB-D approaches, ours performed comparably well in both datasets and even outperformed NOCS~\cite{wang2019normalized} on the 3D IoU metrics in the CAMERA25 dataset, and on the $10^\circ$ metric in the REAL275 dataset.
Also, \figref{fig:syn_sota_rgbd_fig} shows that our method achieved higher performance after around $12^\circ$ threshold errors than NOCS~\cite{wang2019normalized} on the mAP of the rotation metric.
The results show that our proposed algorithm was powerful in the RGB-only comparison, and was also comparable to previous RGB-D methods while using only a single pixel depth value without any additional training.

\subsubsection{Qualitative results}
\figref{fig:REAL275_qualitative} and \figref{fig:qualitative} show examples of our predictions on both the REAL275 and CAMERA25 datasets.
The REAL275 consists of 6 scenarios and \figref{fig:REAL275_qualitative} visualizes each scene of the 3D shape and 6D pose results.
It shows that our method estimated the 3D shape and 6D pose pretty well, even when the appearance and the position and orientation were different.
\figref{fig:qualitative} shows that our method jointly predicted the object shape with fine details such as a bottle head, a mug, and a thin laptop with the accurate pose.
Our object prediction reconstructed well, even for invisible parts.

%%%%%%%%%%%%%%%%%%%%%%%%%%%%%%%%%%%%%%%%%%%%%%%%%%%%%%%%%%%%%%%%%%%%%%%%%%%%%%%%%
%%%%%%%%%%%%%%%%%%%%%%%%%%%%%%%% Conclusion %%%%%%%%%%%%%%%%%%%%%%%%%%%%%%%%%%%%%
%%%%%%%%%%%%%%%%%%%%%%%%%%%%%%%%%%%%%%%%%%%%%%%%%%%%%%%%%%%%%%%%%%%%%%%%%%%%%%%%%

\section{CONCLUSIONS}
In this paper, we proposed a framework for estimating the metric scale shape and 6D pose of a category-level object utilizing only a single RGB image with none or very spare depth information.
Our framework consists of two branches, where the Metric Scale Object Shape branch (MSOS) predicts the metric scale object mesh and the Normalized Object Coordinate Space branch (NOCS) estimates the normalized object coordinate space map.
In our MSOS branch, we propose the Normalized Object Center Estimation (NOCE) with a shared feature encoder to implicitly guide the geometrically aligned object center prediction using 3D volume information.
Using additional similarity transformation between the predicted metric scale mesh and NOCS map, our algorithm outputs metric scale mesh, 6d pose, and the size of an object.
Our results show that our method achieved the state-of-the-art performance for RGB-based methods and our RGB-OD approach was on par and sometimes better performance than the RGB-D methods.

\section*{Acknowledgment}
We would like to thank Kiru Park for valuable comments and insightful discussions.

{\small
\bibliographystyle{ieee_fullname}
\bibliography{egbib}
}

\end{document}